%% file: main.tex
\documentclass[10pt,twocolumn,letterpaper]{article}

\usepackage[pagenumbers]{cvpr} % To force page numbers, e.g. for an arXiv version

\usepackage{xspace}
\usepackage[verbose]{placeins}
\usepackage{xcolor,colortbl,tabularx} 
\usepackage{tikz}
\usepackage{multicol,multirow}
\usepackage{microtype}
\usepackage{tcolorbox}
\usepackage[frozencache,cachedir=minted-cache]{minted}
\definecolor{LightGray}{gray}{0.9}

\newenvironment{tight_itemize}{
\begin{itemize}[leftmargin=15pt,nosep]
  \setlength{\topsep}{0pt}
  \setlength{\itemsep}{0pt}
  \setlength{\parskip}{0pt}
  \setlength{\parsep}{0pt}
}{\end{itemize}}

\newcommand{\llmassist}{\textsc{LLM-Assist}\xspace}
\newcommand{\llmdrive}{\ensuremath{\textsc{LLM-Assist}_{\textsc{unc}}}\xspace} 
\newcommand{\llmhybrid}{\ensuremath{\textsc{LLM-Assist}_{\textsc{par}}}\xspace} 
\newcommand{\gptthreeunc}{\ensuremath{\textsc{GPT-3-Assist}_{\textsc{unc}}}\xspace}
\newcommand{\gptthreepar}{\ensuremath{\textsc{GPT-3-Assist}_{\textsc{par}}}\xspace}
\newcommand{\gptfourunc}{\ensuremath{\textsc{GPT-4-Assist}_{\textsc{unc}}}\xspace}
\newcommand{\gptfourpar}{\ensuremath{\textsc{GPT-4-Assist}_{\textsc{par}}}\xspace}

\definecolor{cvprblue}{rgb}{0.21,0.49,0.74}
\usepackage[pagebackref,breaklinks,colorlinks,citecolor=cvprblue]{hyperref}

\def\paperID{****} % *** Enter the Paper ID here
\def\confName{CVPR}
\def\confYear{2024}

\title{\llmassist: Enhancing Closed-Loop Planning with Language-Based Reasoning}
\author{S P Sharan$^1$ \quad
Francesco Pittaluga$^2$
\quad
Vijay Kumar B G$^2$ 
\quad
Manmohan Chandraker$^{2,3}$\\\\
$^1$ UT Austin \quad $^2$ NEC Labs America \quad $^3$UC San Diego
}

\begin{document}
\maketitle
\input{sec/0_abstract}    
\input{sec/1_intro}
\input{sec/2_related_work}
\input{sec/3_method}
\input{sec/4_exp}
\input{sec/5_conc}
{
    \small
    \bibliographystyle{ieeenat_fullname}
    \bibliography{main}
}

\input{sec/6_suppl}

\end{document}

%% file: sec/0_abstract.tex
\begin{abstract}
    Although planning is a crucial component of the autonomous driving stack, researchers have yet to develop robust planning algorithms that are capable of safely handling the diverse range of possible driving scenarios. Learning-based planners suffer from overfitting and poor long-tail performance \cite{zhai2023ADMLP}. On the other hand, rule-based planners generalize well, but might fail to handle scenarios that require complex driving maneuvers \cite{Dauner2023CORL}. To address these limitations, we investigate the possibility of leveraging the common-sense reasoning capabilities of Large Language Models (LLMs) such as GPT4 \cite{openai2023gpt} and Llama2 \cite{touvron2023llama2} to generate plans for self-driving vehicles. In particular, we develop a novel hybrid planner that leverages a conventional rule-based planner in conjunction with an LLM-based planner. Guided by commonsense reasoning abilities of LLMs, our approach navigates complex scenarios which existing planners struggle with, produces well-reasoned outputs while also remaining grounded through working alongside the rule-based approach. Through extensive evaluation on the nuPlan benchmark, we achieve state-of-the-art performance, outperforming all existing pure learning- and rule-based methods across most metrics. 
    Our code will be available at \href{https://llmassist.github.io}{https://llmassist.github.io}.
\end{abstract}

%% file: sec/1_intro.tex
\vspace{-1em}
\section{Introduction}
\label{sec:intro}

In recent years, aided by advances in deep learning, novel sensing technologies, and low-cost graphics processing units (GPUs), self-driving vehicles have taken major leaps forward. We've even witnessed the deployment of fully self-driving taxi services in limited areas of certain cities. That said, developing planning algorithms for self-driving vehicles capable of handling all the complexities of driving in fully unconstrained environments still remains a significant challenge. 

While deep learning has had major impacts on the perception and prediction components of the self-driving stack, it has yet to have a major impact on closed-loop planning. This is evidence by the fact that a rule-based planning algorithm \cite{Dauner2023CORL} just won the nuPlan benchmark competition at CVPR 2023 \cite{caesar2021nuplan} for both the close-loop non-reactive and reactive settings. A possible challenge for learning-based planners might be that their training in an open-loop setting fails to generalize to a closed loop setting, while training in a closed-loop setting fails to converge to reasonable solutions. On the other hand, while rule-based planners can succeed in most settings, they are not scalable, as it's not possible to enumerate all possible driving scenarios.

The question we seek to answer in this paper is: Can we leverage the common-sense reasoning of LLMs to overcome the limitations of existing learning- and rule-based planners? We answer this in the affirmative, through the key insight that judicious use of an LLM can supplement an existing base planner to perform well in conditions where it might otherwise suffer. Our base planner, PDM-Closed, achieved the previous SOTA on the nuPlan benchmark, through an intelligent driver model-based approach that controls centerline offsets and target speeds. First, we define conditions based on scored proposals of the base planner where the LLM is automatically invoked. Next, we consider an unconstrained LLM planner that must directly return a safe future trajectory, which works surprisingly well but falls short of the base planner in closed loop evaluation. Finally, we allow the LLM to define the planner parameters to safely navigate a scenario, which we find can significantly overcome deficiencies in the base planner. Distinct from prior works that directly use LLMs for planning, we believe this is the first demonstration of an LLM controlling an existing planner to supplement it. 

\begin{figure*}
    \centering
    \includegraphics[width=1\linewidth]{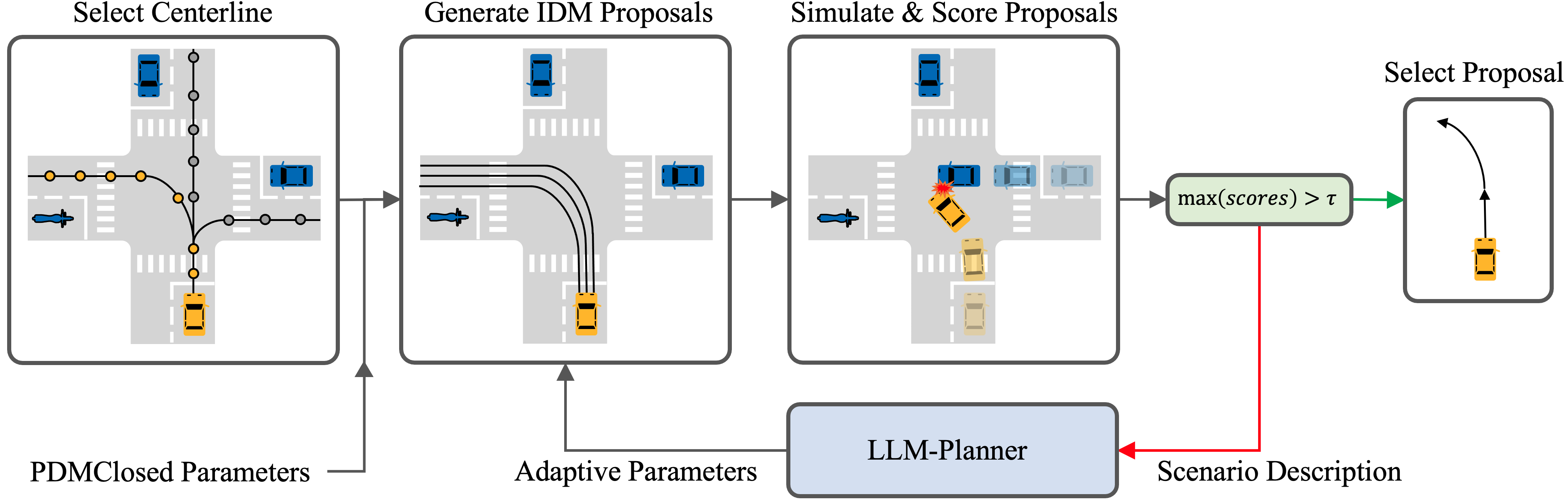}
    \caption{\textbf{Architecture of \llmassist.} We propose a novel hybrid planning approach that leverages a SoTA rule-based planner, PDM-Closed, for common scenarios and a novel LLM-based planner, for challenging high uncertainty scenarios. When the PDM-generated scores are deemed insufficient—falling below predetermined thresholds for various metrics such as collision risk and passenger comfort—we invoke the LLM-Planner.}
    \label{fig:overview}
    \vspace{-1em}
\end{figure*}

In extensive experiments on the nuPlan benchmark, our LLM-assisted planner achieves SOTA performance on the nuPlan benchmark in both the challenging closed-loop reactive and non-reactive settings. We demonstrate several qualitative examples where the LLM-assisted planner can perform non-trivial maneuvers to navigate complex scenarios where the base planner does not succeed or achieves suboptimal safety, efficiency or comfort outcomes. Importantly, the use of an LLM also allows obtaining reasoned outputs for the planner behavior, where our approach of controlling physically meaningful planning variables leads to a well-grounding reasoning.

Our contributions can be summarized as follows:
\begin{tight_itemize}
    \item A strategy to invoke an LLM based on the scores of a base planner, which allows us to judiciously exploit the strengths of both the rule-based planner and the LLM.
    \item An unconstrained hybrid rule- and LLM-based planner that processes text-based scene descriptions as input to directly generate a navigational plan for the ego vehicle as 2D coordinates.
    \item  A hybrid planner that processes text-based scene descriptions to provide parameters for a base planner, PDM-Closed, to plan a safe trajectory for the ego car.
\end{tight_itemize}

%% file: sec/2_related_work.tex
\section{Related Work}
\label{sec:related}

\subsection{Planning for Autonomous Driving}
    Rule-based planning involves the use of explicit rules to guide the decision-making process in autonomous vehicles \cite{thrun2006stanley,bacha2008odin,leonard2008perception,urmson2008autonomous,chen2015deepdriving,fan2018baidu}. It provides a structured and easily understandable framework for determining how the vehicle should behave in different situations. One well-known example of rule-based planning is the Intelligent Driver Model (IDM) \cite{treiber2000congested}, which is designed to help vehicles follow a leading vehicle while maintaining a safe following distance. PDM-Closed \cite{Dauner2023CORL}, an extension of IDM, which executes multiple policies with different settings and evaluates them to choose the best course of action, recently won the nuPlan benchmark challenge \cite{caesar2021nuplan}.
    
    In addition to pure rule-based planning, previous research has explored hybrid approaches that combine rule-based decision-making with machine learning components \cite{Dauner2023CORL,huang2023gameformer,hu2023planning,chen2023tree,rhinehart2021contingencies,song2020pip,hu2022st,wei2021perceive,sadat2020perceive,chen2015deepdriving,cui2021lookout,zeng2019end,chekroun2023mbappe}. These hybrid planners often involve forecasting future environmental conditions, which allows for informed and adaptable driving decisions. This forecasting can take different forms, such as agent-centric, where trajectories are predicted for each vehicle in the environment, or environment-centric, which involves occupancy or cost maps. Furthermore, the forecasting can be influenced by the ego vehicle's plan, taking into account how the ego vehicle's actions affect the future of the scene.
    
\subsection{Large-Language Models}
    Large Language Models (LLMs) like GPT \cite{brown2020language}, its successors GPT-3 and GPT4 \cite{openai2023gpt}, and its open-source counterparts Llama \cite{touvron2023llama} and Llama2 \cite{touvron2023llama2} are a type of artificial intelligence that are designed to understand, generate, and manipulate human language. Built on advanced machine learning algorithms, particularly deep neural networks, these models undergo extensive training on extensive text datasets. This training enables them to grasp the intricacies of language, encompassing grammar, syntax, and semantics. InstructGPT \cite{ouyang2022training}, a specialized version of OpenAI's GPT models, is fine-tuned for superior adherence to user instructions, offering more precise and context-relevant responses across various applications like content creation and information retrieval. ``Chain-of-Thought'' reasoning \cite{wei2022chain} introduces a novel model for large language models, enhancing their complex reasoning capabilities through a series of intermediate reasoning steps, significantly improving performance on tasks involving arithmetic, commonsense, and symbolic reasoning. ``ReAct'' \cite{yao2022react} introduces a new paradigm that enhances the capabilities of large language models in complex tasks requiring both reasoning and decision-making, by prompting these models to generate verbal reasoning traces and actions in an interleaved manner, enabling dynamic reasoning and interaction with external environments.

\subsection{Planning with Large-Language Models}
    Given their common-sense reasoning capabilities, there have been some efforts to leverage LLMs for planning tasks in robotics. A recent approach \cite{ahn2022can}, allows robots to understand and execute high-level textual instructions for physically grounded tasks, merging pre-trained skills with language model insights to ensure feasible and contextually relevant actions. Another work \cite{huang2022inner}, proposed the use of environment feedback to form an inner monologue, enhancing planning and interaction in robotics by integrating perception models and pre-trained skills for improved completion of complex, long-horizon tasks. Similarly, it was shown in \cite{song2023llm} that enhancing LLMs with physical grounding, allowing them to generate and update plans that are contextually relevant to the current environment. 

%% file: sec/3_method.tex
\input{tbl/prompt_unc}
\input{tbl/prompt_param}

\section{Method}
    We propose a novel hybrid planning approach that leverages a SoTA rule-based planner, PDM-Closed, for common scenarios and a novel LLM-based planner, for challenging high uncertainty scenarios. The PDM algorithm is integral to our method, generating 15 trajectory proposals at each planning stage, each characterized by varying velocities and center-lane offsets. These proposals are subsequently assessed using an internal simulator, which applies metrics analogous to those used in the nuPlan Challenge. Our methodology builds upon this framework. When the PDM-generated scores are deemed insufficient—falling below predetermined thresholds for various metrics such as collision risk and passenger comfort—we invoke the LLM-Planner. The LLM-Planner processes the current scenario's observations, including vehicular positioning, traffic light statuses, and lane information, along with the PDM-generated trajectories and their corresponding scores. We propose two LLM-based planners. The first, \llmdrive considers the most unconstrained version of the planning problem, in which the LLM must directly return a safe future trajectory for the ego car. The second, \llmhybrid considers a parameterized version of the planning problem, in which the LLM must only return a set of parameters for a rule-based planner, PDM-Closed \cite{Dauner2023CORL}.

    \begin{figure}
        \centering
        \includegraphics[width=\linewidth]{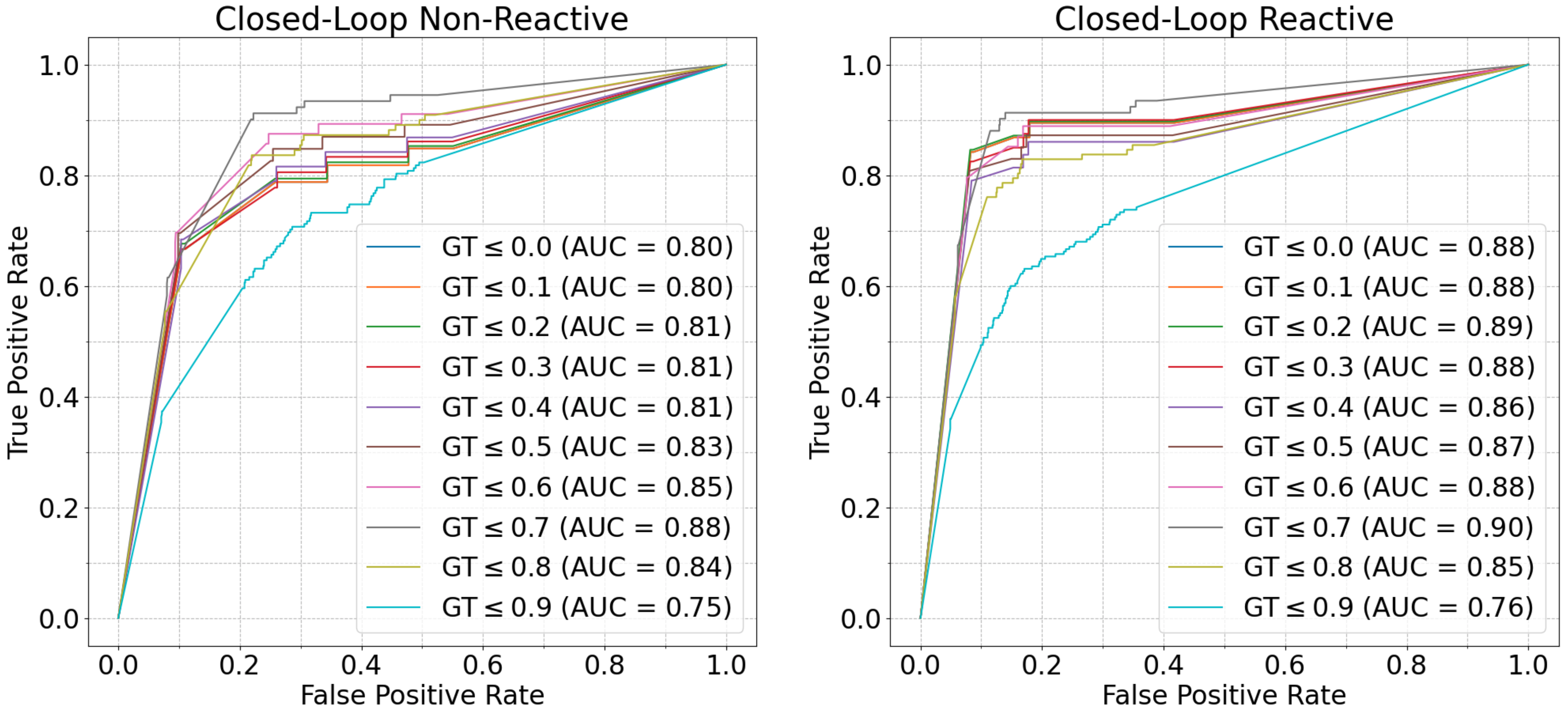}
        \caption{\textbf{ROC Curve of PDM-Closed's Predicted Proposal Scores.} Evaluated on nuPlan Closed-Loop Challenges Val14 split.}
        \label{fig:pdm_classifier}
        \vspace{-1em}
    \end{figure}

    \subsection{Base Planner}
        The strong performances of IDM \cite{treiber2000congested} and PDM-Closed \cite{Dauner2023CORL} on the closed-loop nuPlan evaluations demonstrate that rule-based planners are capable of successfully maneuvering the vast majority of driving scenarios. As such, we propose using a rule-based planner as a base planner and only invoking the LLM-based planner for challenging scenarios that the rule-based planner cannot solve. The challenge, however, is how to identify which scenarios the rule-based planner cannot solve. For this, we leverage a simple constant-velocity real-time simulator to score the proposals from our rule-based planner and invoke the LLM only when the score falls below a predetermined threshold.

        \input{tbl/sweep}
        \input{tbl/res}

        For both the base planner and the real-time simulator, we build on PDM-Closed \cite{Dauner2023CORL}. PDM-Closed \cite{Dauner2023CORL} is a rule-based planner that generates 15 trajectory proposals for the ego vehicle at each time step. For each proposal, a constant velocity simulation that considers all agents within a 50-meter radius is carried out and the top scoring proposal, according to the nuPlan Challenge's metrics \cite{caesar2021nuplan}, is selected. If the top scoring proposal is set to collide within 2 seconds, an emergency brake function is triggered. The 8 metrics  considered in the nuPlan Challenge are \textit{no\_ego\_at\_fault\_collisions}, \textit{drivable\_area\_compliance}, \textit{ego\_is\_making\_progress}, \textit{driving\_direction\_compliance}, \textit{time\_to\_collision\_within\_bound}, \textit{speed\_limit\_compliance}, \textit{ego\_progress\_along\_expert\_route}, \textit{ego\_is\_comfortable}. 

        The 15 trajectories are generated by leveraging the intelligent driver model (IDM) and considering all possible combinations of 2 hyperparameters: centerline offset $o=\{-1, 0, 1\}$ meters and target speed $v_0=\{20\%, 40\%, 60\%, 80\%, 100\% \}$ of the designated speed limit. Specifically, given a centerline offset $o$ and a current velocity $v$, the  longitudinal acceleration for a proposal is generated via the intelligent driver model (IDM)
        \begin{equation}
            \frac{dv}{dt} = a\Bigg(1-\bigg(\frac{v}{v_0}\bigg)^\delta - \bigg(\frac{s^*}{s}\bigg)^2 \Bigg),
            \label{eq:idm}
        \end{equation}
        where $a$ denotes acceleration limit, $s^*$ safety margin, and $\delta$ exponent (or jerk). 
        
    \subsection{\llmdrive}
    \label{subsection_llmdrive}
        \llmdrive considers the most unconstrained version of the planning problem, in which the LLM must directly return a safe future trajectory for the ego car. For this task, we provide the system prompt shown in \Cref{listing:prompt_unc}. The system prompt consists of a task overview, a description of the state of the scene and task requirements composed of generating a trajectory and a rationale. We access the assets provided by the nuPlan API to input the lane geometry and states of all agents, including the ego-vehicle. 

    \subsection{\llmhybrid}
        \llmhybrid considers a parameterized version of the planning problem. Rather than directly returning a future trajectory for the ego car, the LLM instead returns a set of parameters to be used by the base planner, PDM-Closed, to plan a safe trajectory for the ego car. For this task, we provide the system prompt shown in \Cref{listing:prompt_param} at each time step. As in for \llmdrive, we at each time step,  provide scene information including the history of vehicle, pedestrian, and object positions, headings, and speeds, and their current lane ID. The task of the LLM is to return valid values for the following parameters:
      
        \begin{enumerate}
            \item $\texttt{lateral\_offsets}$: Ego offset relative to lane center.
            \item $\texttt{speed\_limit\_fraction}$: Speed-limit fraction in free traffic.
            \item $\texttt{fallback\_target\_velocity}$: Fallback speed in free traffic.
            \item $\texttt{min\_gap\_to\_lead\_agent}$: Min distance to lead car.
            \item $\texttt{headway\_time}$: Min time to the lead car.
            \item $\texttt{accel\_max}$: Max acceleration.
            \item $\texttt{decel\_max}$: Max deceleration.
        \end{enumerate}
        
        Additionally, the LLM should provide a one sentence explanation for why the specific trajectory was chosen. As illustrated in \cref{fig:overview}, \llmassist queries the LLM planner multiple times at a given time step until a trajectory is proposed that has a predicted score that meets a predefined threshold or the number of queries per time step exceeds a predefined threshold. If the query threshold is exceeded, the trajectory with the highest predicted score is selected. 

%% file: tbl/prompt_unc.tex
\begin{listing}
    \begin{tcolorbox}[
        colframe=gray,
        colbacktitle=gray,
        sharp corners,
    ]
        \fontsize{3.5}{4.2} % the second number should always be 1.2x the first
        % \inputminted[breaksymbolleft=]{text}{tbl/system_prompt_unconstrained.txt}
\begin{minted}{text}
# Task Overview
You are the decision-making agent for a self-driving car in a simulated 2D top-down environment. Your objective is to plan the car's trajectory, ensuring safe navigation based on the dynamic positions and speeds of surrounding vehicles and objects.

# Simulation Environment
- **Perspective**: Top-down view in 2D space. The scene is centered with the ego vehicle at (0,0).
- **Update Frequency**: Each conversation turn provides an updated scenario.

# Inputs
- **Historical Data**: Position, heading, speed, lane data for the ego vehicle and other entities (other vehicles, traffic cones, etc.) are provided for three time points: t-2, t-1, and t (present). Heading is in radians with 0 being the positive x-axis and pi/2 being the positive y-axis.
- **Data Format**:
  - **Ego Vehicle**: Position (list of tuples), heading (tuple), speed (tuple), lane.
  - **Other Entities**: Similar data structure for position, heading, speed, and lane.

Example Scenario Description:
- Ego Vehicle
	- Position: ([-13.28, 10.29], [-6.6, 5.27], [0.0, 0.0])
	- Heading: (-0.62, -0.66, -0.68)
	- Speed: (8.24, 8.66, 8.63)
    - Lane: 18553
- [Other Entities]

# Task Requirements
Generate the following and nothing else:
1. **Trajectory Planning**:
   - Create an 8-second trajectory for the ego car using historical data.
   - Utilize lane information to understand where each vehicle is positioned. Follow the lane you are in and do not change lanes unless necessary (collision scenarios).
   - Generate 4 waypoints at 2-second intervals (t + 2, t + 4, t + 6, t + 8). Only generate these future waypoints and do not repeat the ego history.
   - Each waypoint is an (x, y) coordinate.
   - Format: "Trajectory: [(x_2sec, y_2sec), (x_4sec, y_4sec), (x_6sec, y_6sec), (x_8sec, y_8sec)]".

2. **Decision Rationale**:
   - Provide a short one sentence brief explanation for your trajectory choice.
   - Format: "Reasoning: [Your reasoning]".

Again, do not deviate from this output format. An example is shown below.

Example Output:
Trajectory: [(2.13, -1.83), (4.26, -3.66), (6.39, -5.49), (8.52, -7.32)]
Reasoning: The trajectory maintains the current lane and speed while avoiding nearby vehicles, which are either stationary or moving away.

# Objective
- Ensure the safety of the ego car by avoiding collisions.
- Adapt the trajectory based on the historical movement patterns of surrounding entities.
- Prioritize realistic and feasible routes in the given dynamic environment.
\end{minted}
    \end{tcolorbox}
    \vspace{-1em}
    \caption{\textbf{System Prompt for \llmdrive.}\vspace{-1em}}
    \label{listing:prompt_unc}    
\end{listing}

%% file: tbl/prompt_param.tex
\begin{listing}
    \begin{tcolorbox}[
        colframe=gray,
        colbacktitle=gray,
        sharp corners,
    ]
        % \tiny
        \fontsize{3.5}{4.2} % the second number should always be 1.2x the first
        % \inputminted[breaksymbolleft=]{text}{tbl/system_prompt_constrained.txt}
\begin{minted}{text}
# Task Overview
You are the decision-making agent for a self-driving car in a simulated 2D top-down environment. Your objective is to select parameters to be used by a planner to plan the car's trajectory, ensuring safe navigation based on the dynamic positions and speeds of surrounding vehicles and objects. Do not change the default parameters unless necessary.

# Parameters that can be changed
Here are different parameters:
lateral_offsets: The offset of the ego car relative to the centerline of the current lane. Negative values move the ego car to the right and positive values to the left. For example, an offset of -2.5 moves the ego car to the left lane.
speed_limit_fraction: Fraction of the speed-limit desired in free traffic. You can choose float values between 0 and 1, where 0 indicates the slowest speed and 1 indicates the fastest.
fallback_target_velocity: This is the desired fallback velocity in free traffic in meters per second. The default value is 15.
min_gap_to_lead_agent: Minimum relative distance to lead car in meters. The default value is 1.
headway_time: This is the desired time headway. The minimum time to the lead car in seconds. The default value is 1.5.
accel_max: Maximum acceleration in meters per second square. The default value is 5.
decel_max: Maximum deceleration (positive value) in meters per second squared. The default value 3.
emergency_brake: This is a float value that indicates whether the ego car should brake in case of emergency. Default value is 0.0, if you want the ego car to brake in case of emergency, then set this value to 1.0

These parameters are organized as follows:
```python
# This is the `new_args` dictionary whose values you will update and values shown below are the default values: 
new_args = {
    "lateral_offsets": [-1.0, 0.0, 1.0],
    "fallback_target_velocity": 15.0,
    "speed_limit_fraction": [0.2, 0.4, 0.6, 0.8, 1.0],
    "min_gap_to_lead_agent": 1.0,
    "headway_time": 1.5,
    "accel_max": 5.0,
    "decel_max": 3.0,
	"emergency_brake": 0.0,
}
```

# Simulation Environment
- **Perspective**: Top-down view in 2D space. The scene is centered with the ego vehicle at (0,0).
- **Update Frequency**: Each conversation turn provides an updated scenario.

# Inputs
- **Predicted Scores**: The predicted scores for the parameter combinations provided. The scores are in the range of 0 to 1, with 1 being the highest score. The higher the score, the safer the parameters are for the current scenario. The scores and their corresponding parameters are provided as a comma seperated list.
- **Historical Data**: Position, heading, speed, lane data for the ego vehicle and other entities (other vehicles, traffic cones, etc.) are provided for three time points: t-2, t-1, and t (present). Heading is in radians, with 0 being the positive x-axis and pi/2 being the positive y-axis.
- **Data Format**:
  - **Ego Vehicle**: Position (list of tuples), heading (tuple), speed (tuple), lane.
  - **Other Entities**: Similar data structure for position, heading, speed, and lane.

Example:
- predicted_score, lateral_offsets, fallback_target_velocity, speed_limit_fraction, min_gap_to_lead_agent, headway_time, accel_max, decel_max, emergency_brake
	- 0.3, -1.0, 15.0, 0.2, 1.0, 1.5, 5.0, 3.0, 0.0
	- 0.1, 0.0, 1.0, 15.0, 0.6, 1.0, 1.5, 5.0, 3.0, 0.0
	- [Other Scores]
- Ego Vehicle
	- Position: ([-13.28, 10.29], [-6.6, 5.27], [0.0, 0.0])
	- Heading: (-0.62, -0.66, -0.68)
	- Speed: (8.24, 8.66, 8.63)
    - Lane: 18553
- [Other Entities]

# Task Requirements
Always generate the following two outputs (1. Python Code Block and 2. Decision Rationale) and nothing else:
1. **Python Code block**
- In A python code block, propose only the parameters that you want to change. For example:
```python
new_args["lateral_offsets"] = [2.5]
new_args["speed_limit_fraction"] = np.array([0.1, 0.2])
```
- Keep the variable names as they are. Apart from `lateral_offsets`, the other variables are float NumPy arrays. Do not change the parameters unless necessary. It's ok to not propose any changes if everything looks safe.

2. **Decision Rationale**:
- Provide a short one sentence brief explanation for your parameter choice.
- Format: "Reasoning: [Your reasoning]".

# Tips for Good Proposal
- Always include decision rationale in response.
- Ensure the safety of the ego car by avoiding collisions.
- Adapt the trajectory based on the historical movement patterns of surrounding entities.
- Prioritize realistic and feasible routes in the given dynamic environment.
- Do not change the default parameters unless necessary. 
- It's ok to not propose any changes if everything looks safe.
- Do not propose changes that can cause collisions. 
- Do not propose manuers that can cause the ego car to go close to other vehicles. 
- Try to be away by at least six meters from the vehicle in front of you.
- You will be rewarded for the following:
	- avoiding collision with other vehicles and objects
	- keeping the default parameters whenever it is safe to do so
	- Staying close to the centerline
\end{minted}
    \end{tcolorbox}
    \vspace{-1em}
    \caption{\textbf{System Prompt for \llmhybrid.}\vspace{-1em}}
    \label{listing:prompt_param}
\end{listing}

%% file: tbl/sweep.tex
\begin{table*}
\centering
\setlength{\aboverulesep}{0pt}
\setlength{\belowrulesep}{0pt}
\setlength{\extrarowheight}{.75ex}
% \resizebox{\linewidth}{!}{%
\begin{tabular}{>{\columncolor{gray!50}}c >{\columncolor{gray!15}}c ccccccccc}
\toprule
\textbf{\# Proposals} & \textbf{Score} & \textbf{Collisions} & \textbf{TTC}  & \textbf{Drivable}  & \textbf{Comfort}  & \textbf{Progress}  & \textbf{Speed Limit}  & \textbf{Direction} \\
\midrule
15 & \textbf{92.51} & \textbf{98.05} & \textbf{93.11} & \textbf{99.55} & \textbf{95.19} & 91.75 & \textbf{99.83} & \textbf{99.95} \\
8505 & 77.78 & 91.92 & 62.89 & 98.64 & 78.68 & \textbf{95.60}  & 99.78 & 99.36\\
\bottomrule
\end{tabular}
% }
\smallskip
\caption{\textbf{Ablation of Number of PDMClosed Proposals.} PDMClosed evaluated nuPlan Closed-Loop Non-Reactive Challenge on Val14 split. PDMClosed fails to select the best proposal when presented with too many options, as it relies on a constant velocity simulator.}
\label{tab:sweep}
\end{table*}

%% file: tbl/res.tex
\begin{table*}
\centering
\setlength{\aboverulesep}{0pt}
\setlength{\belowrulesep}{0pt}
\setlength{\extrarowheight}{.75ex}
\resizebox{\linewidth}{!}{%
\begin{tabular}{c >{\columncolor{gray!50}}l >{\columncolor{gray!15}}c ccccccccc}
\toprule
\textbf{Challenge} & \textbf{Method} & \textbf{Score} & \textbf{Collisions} & \textbf{TTC}  & \textbf{Drivable}  & \textbf{Comfort}  & \textbf{Progress}  & \textbf{Speed Limit}  & \textbf{Direction} \\
\midrule
\multirow{3}{*}{\shortstack{Closed-Loop\\Non-Reactive}} 
& PDMClosed  & 92.51 & 98.05 & 93.11 & \textbf{99.55} & 95.19 & 91.75 & \textbf{99.83} & \textbf{99.95} \\
& \gptthreeunc & 90.11 & 96.19 & 92.55 & 98.91 & 93.37 & 91.05 & \textbf{99.83} & 99.91 \\
& \gptthreepar & \textbf{93.05} & \textbf{98.31} & \textbf{93.69} & 99.54 & \textbf{95.61} & \textbf{92.16} & \textbf{99.83} & \textbf{99.95} \\
\midrule
\multirow{3}{*}{\shortstack{Closed-Loop\\Reactive}} 
& PDMClosed  & 91.79 & 97.91 & 93.29 & 99.37 & 94.65 & 89.92 & \textbf{99.83} & \textbf{99.95} \\
& \gptthreeunc & 90.32 & 96.82 & 93.10 & 98.73 & 92.92 & 89.01 & \textbf{99.83} & 99.86 \\
& \gptthreepar & \textbf{92.20} & \textbf{98.18} & \textbf{93.62} & \textbf{99.64} & \textbf{94.72} & \textbf{90.07} & \textbf{99.83} & \textbf{99.95} \\
\bottomrule
\end{tabular}
}
\smallskip
\caption{\textbf{\llmassist evaluated on nuPlan Closed-Loop Challenges on Val14 split.} \gptthreepar achieves SoTA performance on almost all metrics on both closed-loop challenges, reducing the number of dangerous driving scenarios by 11\%.}
\label{tab:nuplan}
\end{table*}

%% file: sec/4_exp.tex
\input{tbl/gpt3}
\input{tbl/brake}
\input{tbl/temp}
\input{tbl/gpt3vs4}

\section{Experimental Setup}

    \paragraph{nuPlan Benchmark.} The nuPlan benchmark \cite{caesar2021nuplan} is the world's first large-scale planning benchmark for autonomous driving. In addition to releasing 1200 hours of annotated human driving data from 3 cities across the US and Asia, the benchmark outlines three planning challenges: open-loop, closed-loop non-reactive, and closed-loop reactive. In this paper, we focus on the two closed-loop challenges, as they more accurately reflect real-world driving \cite{Dauner2023CORL}. We follow \cite{Dauner2023CORL} and consider the \texttt{val14} subset of the nuPlan benchmark \cite{caesar2021nuplan}. It consists of $100$  scenarios of $14$ scenario types, totaling 1,114 scenarios.  
    
    \vspace{-1em}

    \paragraph{Metrics.} Each column of \Cref{tab:nuplan}, except for the first, shows a different binary metric for some aspect of driving. The scores represent the percentage of scenarios which a vehicle successfully navigated without violating the given constraint. Please see the supplementary material for the full details of each metric.
    
    \vspace{-1em}
    
    \paragraph{Baseline Method.} The score achieved by the base planner, i.e., PDM-Closed is shown in \Cref{tab:nuplan}. As we can observe from \Cref{tab:nuplan}, PDM-Closed is capable of successfully navigating the vast majority of scenarios in the nuPlan dataset. Our goal in this paper is investigating how to integrate an LLM-Planner into the mix to handle the safety critical scenarios that PDM-Closed fails at. 

\section{Results}
    \subsection{Hyperparameter Search}
        The base planner, i.e., PDM-Closed \cite{Dauner2023CORL} creates 15 trajectory proposals per time step for the ego vehicle, comprising combinations of 5 speed limit fractions and 3 lateral offsets. These trajectories are then evaluated using a constant velocity model. The choice of 15 proposals was to reduce  the computational overhead for subsequent stages. However, to test whether a broader range of hyperparameters could enhance the planner's performance, we conducted an experiment using varied hyperparameters: 6 lateral offsets, 3 fallback target velocities, 5 speed limit fractions, 3 minimum gaps to the lead agent, 3 headway times, 3 maximum accelerations, and 3 maximum decelerations which resulted in 8505 trajectory proposals at each time step. The results in \Cref{tab:sweep} indicate that this extensive search for optimal hyperparameters diminished the effectiveness of the PDM planner. We hypothesize that this decline in performance could be linked to the reliance on constant velocity assumptions in the trajectory evaluation process. This suggests that a larger hyperparameter search space does not lead to performance improvement for the base planner. In fact, it harms performance while also worsening planning speed. 
    
    \subsection{Prediction}
        We evaluate the efficacy of the base planner's internal simulator at predicting whether a given trajectory will score highly according to the nuPlan benchmark, which we use as a proxy for successfully navigating a scene. We validate the accuracy of these predicted scores on the val14 subset \cite{Dauner2023CORL} of nuPlan closed-loop reactive benchmark \cite{caesar2021nuplan} by comparing the true average score achieved by PDM-Closed for each scenario to the respective predicted scores. Note, however, since we are interested in identifying the exact time step where the LLM should be invoked, we compare the minimum predicted score over all 150 time steps in a scenario to the ground truth score for the whole scenario. In \Cref{fig:pdm_classifier}, we show ROC curves for predicting whether the ground-truth score is less than a given threshold. The results show that PDM-Closed's internal simulation can relatively reliably predict when it will score poorly on a given scenario. This motivates the design of our combined rule-based and LLM-based planner. If the LLM-based planner can succeed when PDM-Closed predicts that it will score poorly, the result will be a superior planner capable of handling a much wider range of driving scenarios.
    
    \subsection{\llmassist Evaluation}
        As shown in \Cref{tab:nuplan}, \gptthreepar achieves SoTA performance on almost all metrics across both nuPlan Closed-Loop Challenges \cite{caesar2021nuplan} on the Val14 Split \cite{Dauner2023CORL}. Regarding safety, \gptthreepar reduces the number of dangerous driving events by 11\% relative to PDMClosed, the current SoTA. We define a dangerous driving event as any scenario in which the ego vehicle is involved in an actual or near collision, drives off the road, exceeds a safe limit for acceleration or jerk, or drives in the wrong direction. Comparing \gptthreeunc to \gptthreepar highlight the importance of our proposed approach -- having the LLM select parameters for a base planner as opposed to directly generating a trajectory for the ego vehicle. For \gptthreepar, the LLM was invoked a max of four times per time step. 

\begin{figure*}
    \centering
    \includegraphics[width=1\linewidth]{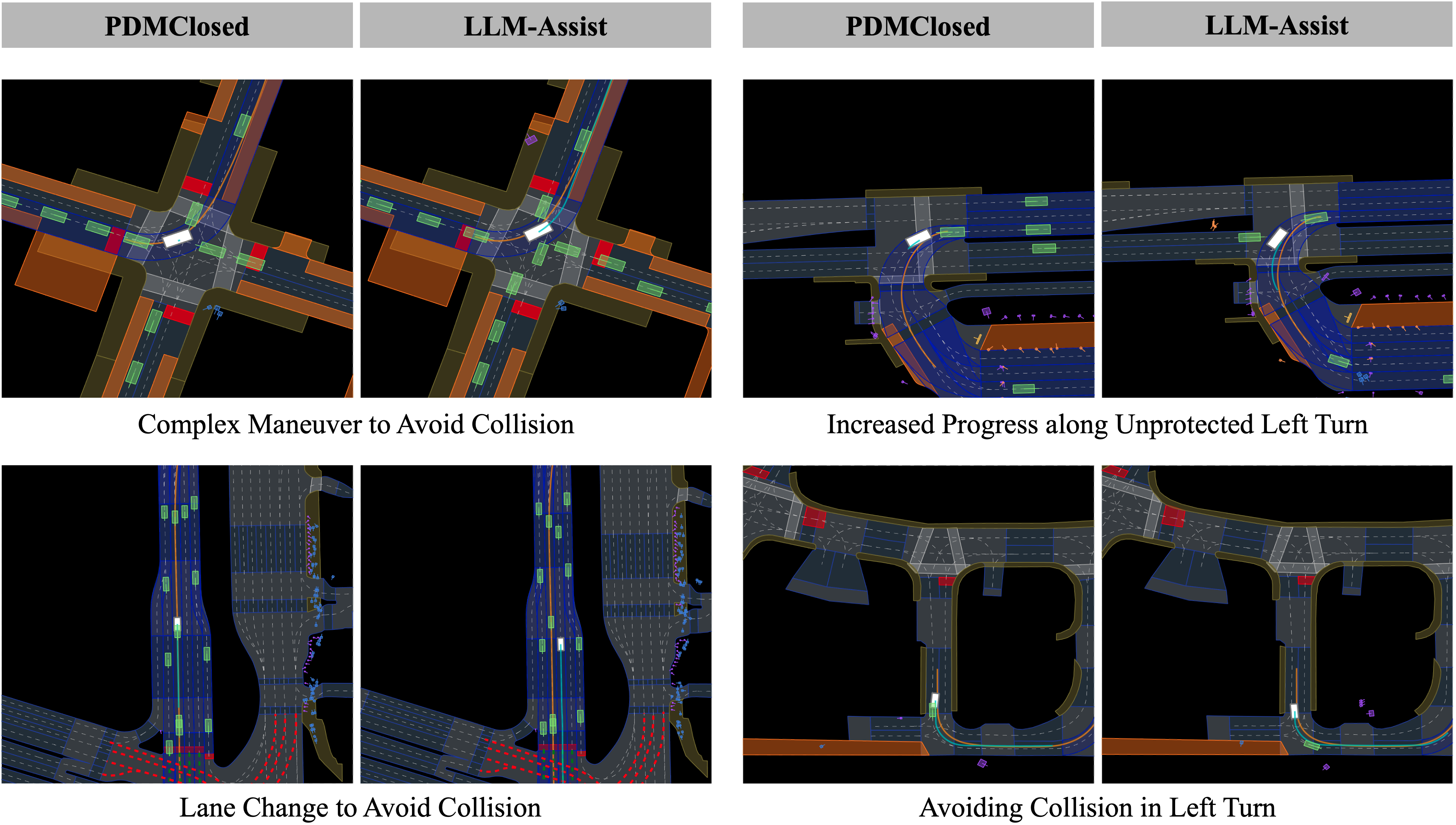}
    \caption{\textbf{Qualitative Comparisons between \llmassist and PDMClosed.}}
    \label{fig_qual}
\end{figure*}

\begin{figure*}
    \centering
    \includegraphics[width=1\linewidth]{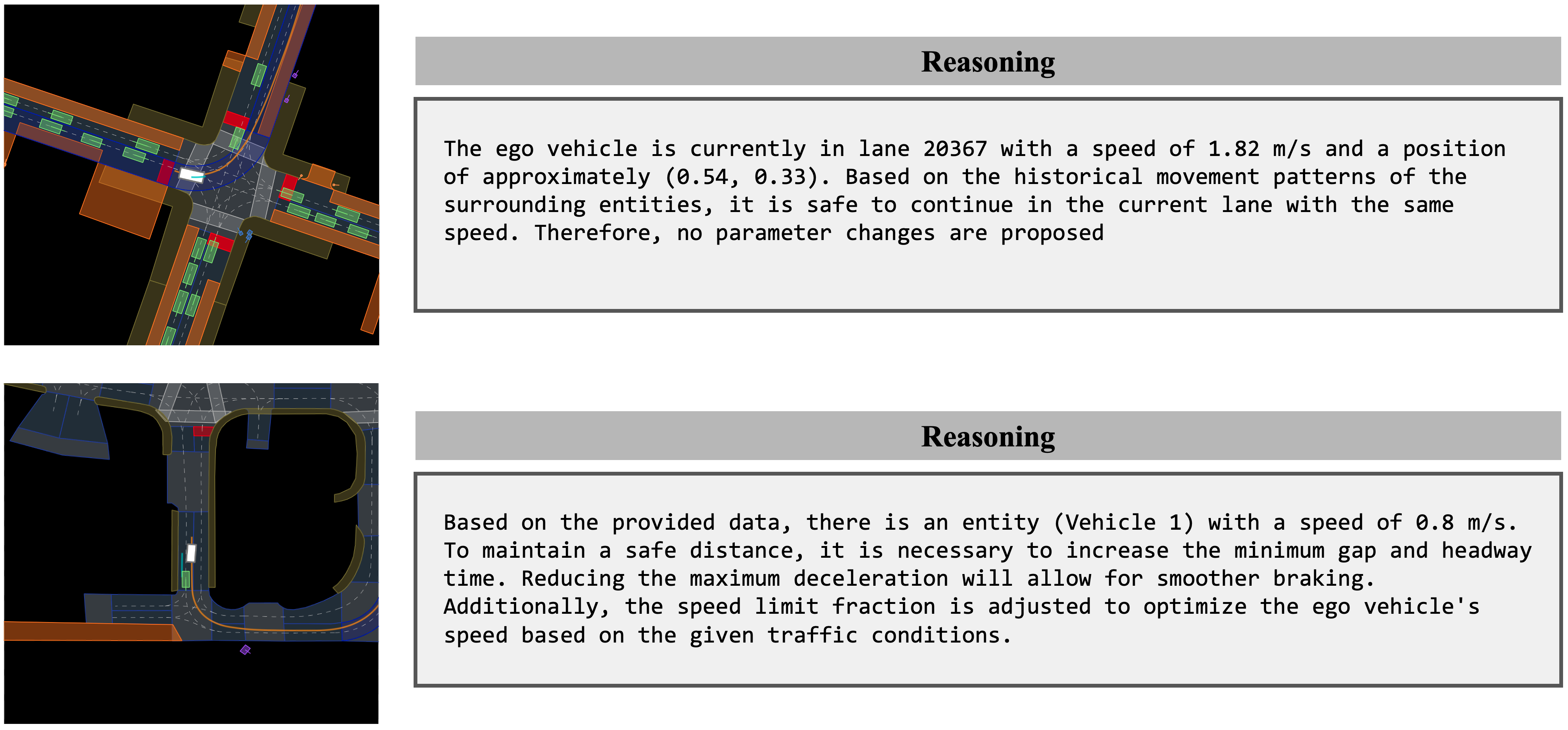}
    \caption{\textbf{Qualitative Example of Reasoning by \llmassist.}}
    \label{fig_qual_reas}
    \vspace{-1em}
\end{figure*}

        In \Cref{fig_qual}, we show qualitative results 
        where \gptthreepar performs complex maneuvers to successfully handle safety-critical scenarios which PDMClosed fails at. In \Cref{fig_qual_reas}, we show interesting reasoning outputs from \gptthreepar. Note, how the reasoning results are grounded by the base planner's parameters. Additional reasoning results are shown in  \Cref{fig_qual_1qnr,fig_qual_1qr,fig_qual_4qnr,fig_qual_4qnr} of the supplementary material. 
    
    \subsection{Ablations}
        For all ablation studies, the LLM was invoked a maximum of one time per time step.
    
        \vspace{-1em}
        
        \paragraph{Importance of Base Planner.} We evaluate a purely LLM-based planner (GPT-3) by providing the system prompt in \Cref{listing:prompt_unc} at each time step in addition to scene information such as the history of vehicle, pedestrian, and object positions, headings, and speeds, and their current lane ID. The LLM needs to provide an 8-second trajectory for the ego car by generating 4 waypoints at 2-second intervals. It's important to note the distinction between this GPT-3 planer and the \llmdrive approach outlined in \Cref{subsection_llmdrive} -- unlike \llmdrive, which is invoked at selected time steps, this planner is invoked at every time step. From \Cref{tab:nuplan-100}, we see that the LLM-based planner (GPT-3) achieves a reasonable performance  $75\%, 73\%, 81\%$ for Collisions, TTC and Drivable metrics respectively. Notably, the GPT-3 achieved this score without explicit training on trajectory data, highlighting its ability to generalize and adapt to driving scenarios. That said, given the low scores relative to \gptthreepar, it's clear why our proposed approach of having the LLM supplement a base planner is a better way to leverage the reasoning capabilities of LLMs for planning.
    
        \vspace{-1em}
    
        \paragraph{Control over Emergency Break.} We explored the impact of granting control over the ``emergency brake" function of PDM-Closed to the LLM. This function, typically activated in anticipation of collisions by PDM's internal simulator, was managed by the LLM to test its efficacy in critical scenarios. As evidenced in our results in \Cref{tab:brake}, this integration significantly enhanced performance metrics. Most notably, it improved collision rates, time to collision, and drivability metrics, demonstrating the LLM's adeptness at judiciously deciding when to activate such crucial safety functionalities. 
    
        \vspace{-1em}
    
        \paragraph{Temperature of LLMs.} We vary the temperature of GPT-3 for \llmdrive and run the nuPlan Closed-Loop Reactive Challenge, and report the results in \Cref{tab:temp}. GPT-3 performs best at a temperature of 1.4, indicating that allowing the LLM greater freedom may lead to better planning.
    
        \vspace{-1em}
        
        \paragraph{LLM Architecture and Timing Analysis.} We perform a comparative analysis of the performance of \llmdrive and \llmhybrid, utilizing both GPT-3 and GPT-4, with detailed results presented in \Cref{tab:gpt3vs4}. The findings reveal that while both GPT-3 and GPT-4 exhibit comparable effectiveness overall, subtle differences were observed: GPT-4 demonstrates a marginal edge in unconstrained settings, whereas GPT-3 shows slightly better performance in constrained ones. Additionally, we investigate the possibility of using an open-source LLM, Llama2-7B \cite{touvron2023llama2}, as the LLM for \llmassist. The result of our investigation was that, without fine-tuning, Llama2-7B is not a suitable LLM for \llmassist, as we were unable to have it reliably produce outputs of any particular format, even with a few in-context examples. Nonetheless, we can still use Llama 2 to get a rough estimate of the run-time for our main results with GPT-3. For a request for a single set of parameters for the planner, Llama 2 takes 3 seconds. 

%% file: tbl/gpt3.tex
\begin{table*}
\centering
\setlength{\aboverulesep}{0pt}
\setlength{\belowrulesep}{0pt}
\setlength{\extrarowheight}{.75ex}
\resizebox{\linewidth}{!}{%
\begin{tabular}{c >{\columncolor{gray!50}}l >{\columncolor{gray!15}}c ccccccccc}
\toprule
\textbf{Challenge} & \textbf{Method} & \textbf{Score} & \textbf{Collisions} & \textbf{TTC}  & \textbf{Drivable}  & \textbf{Comfort}  & \textbf{Progress}  & \textbf{Speed Limit}  & \textbf{Direction} \\
\midrule
\multirow{2}{*}{\shortstack{Closed-Loop\\Non-Reactive}} 
& \textsc{GPT-3} & 18.08 & 63.04 & 60.14 & 78.26 & 57.97 & 31.30 & 99.93 & 98.91 \\
& \gptthreepar  & \textbf{94.80} & \textbf{100.00} & \textbf{94.89 }& \textbf{100.00} & \textbf{97.81} & 90.18 & \textbf{99.86} & \textbf{99.64} \\
\midrule
\multirow{2}{*}{\shortstack{Closed-Loop\\Reactive}} 
& \textsc{GPT-3} & 22.33 & 75.18 & 73.72 & 81.02 & 56.93 & 31.77 & \textbf{99.96} & \textbf{100.00} \\
& \gptthreepar & \textbf{92.82} & \textbf{98.55} & \textbf{97.10} & \textbf{99.28} & \textbf{94.20} & \textbf{89.15} & 99.86 & 99.28 \\
\bottomrule
\end{tabular}
}
\smallskip
\caption{\textbf{GPT-3 vs \llmassist.} Comparison between GPT-3 planner and \llmassist on nuPlan Closed-Loop Challenges on a subset of Val14 split consisting of 140 samples. Without fine-tuning, GPT-3 on its own is incapable of directly generating successful plans. This shows the importance of \llmassist's hybrid architecture.}
\label{tab:nuplan-100}
\end{table*}

%% file: tbl/brake.tex
\begin{table*}[]
\centering
\setlength{\aboverulesep}{0pt}
\setlength{\belowrulesep}{0pt}
\setlength{\extrarowheight}{.75ex}
% \resizebox{\linewidth}{!}{%
\begin{tabular}{>{\columncolor{gray!50}}c >{\columncolor{gray!15}}c ccccccccc}
\toprule
\textbf{Brake} & \textbf{Score} & \textbf{Collisions} & \textbf{TTC}  & \textbf{Drivable}  & \textbf{Comfort}  & \textbf{Progress}  & \textbf{Speed Limit}  & \textbf{Direction} \\
\midrule
$\times$   & 91.85 & 97.81 & 93.99 & 99.36 & \textbf{92.99} & \textbf{90.40} & \textbf{99.83} & \textbf{99.91} \\
\checkmark & \textbf{92.16} & \textbf{97.96} & \textbf{94.10} & \textbf{99.46} & 92.82 & 90.34 & \textbf{99.83} & \textbf{99.91} \\
\bottomrule
\end{tabular}
% }
\smallskip
\caption{\textbf{Ablation Study of Emergency Break.} \gptthreepar evaluated on nuPlan Closed-Loop Reactive Challenge on Val14 split. Enabling the LLM to invoke an emergency brake leads to improved performance, as it can avoid potential collisions.}
\label{tab:brake}
\end{table*}

%% file: tbl/temp.tex
\begin{table*}[]
\centering
\setlength{\aboverulesep}{0pt}
\setlength{\belowrulesep}{0pt}
\setlength{\extrarowheight}{.75ex}
% \resizebox{\linewidth}{!}{%
\begin{tabular}{>{\columncolor{gray!50}}c >{\columncolor{gray!15}}c ccccccccc}
\toprule
\textbf{Temp} & \textbf{Score} & \textbf{Collisions} & \textbf{TTC}  & \textbf{Drivable}  & \textbf{Comfort}  & \textbf{Progress}  & \textbf{Speed Limit}  & \textbf{Direction} \\
\midrule
0.2 & 91.91 & 98.00 & 94.28 & 99.27 & 92.29 & 90.31 & \textbf{99.83} & \textbf{99.95} \\
0.6 & 92.01 & 98.00 & 93.90 & 99.54 & 93.35 & 90.16 & \textbf{99.83} & 99.82 \\
1.0 & 92.16 & 97.96 & \textbf{94.10} & 99.46 & 92.82 & 90.34 & \textbf{99.83} & 99.91 \\
1.2 & 92.21 & \textbf{98.05} & 93.82 & \textbf{99.64} & 93.45 & 90.20 &\textbf{ 99.83} & 99.91 \\
1.4 & \textbf{92.24} & 98.02 & 94.01 & 99.35 & 93.82 & \textbf{90.71} & 99.81 & 99.91 \\
1.6 & 92.14 & 98.04 & 93.72 & 99.36 & \textbf{94.90} & 90.11 & \textbf{99.83} & \textbf{99.95} \\
2.0 & 92.05 & 98.00 & 93.63 & 99.27 & 94.54 & 90.19 & \textbf{99.83} & \textbf{99.95} \\
\bottomrule
\end{tabular}
% }
\smallskip
\caption{\textbf{Ablation Study of GPT-3 Temperature.} \gptthreepar evaluated on nuPlan Closed-Loop Reactive Challenge on Val14 split. \llmassist achieves the best performance with a GPT-3 temperature of 1.4, showing that greater flexibility may allow for better planning.}
\label{tab:temp}
\end{table*}

%% file: tbl/gpt3vs4.tex
\begin{table*}[]
\centering
\setlength{\aboverulesep}{0pt}
\setlength{\belowrulesep}{0pt}
\setlength{\extrarowheight}{.75ex}
% \resizebox{\linewidth}{!}{%
\begin{tabular}{>{\columncolor{gray!50}}c >{\columncolor{gray!15}}c ccccccccc}
\toprule
\textbf{Architecture} & \textbf{Score} & \textbf{Collisions} & \textbf{TTC}  & \textbf{Drivable}  & \textbf{Comfort}  & \textbf{Progress}  & \textbf{Speed Limit}  & \textbf{Direction} \\
\midrule
\gptthreeunc & 90.32 & \textbf{96.82} & 93.10 & \textbf{98.73} & 92.92 & 89.01 & \textbf{99.83} & 99.86 \\
\gptfourunc & \textbf{90.46} & 96.68 & 93.19 & \textbf{98.73} & \textbf{94.55} & \textbf{89.32} & \textbf{99.83} & \textbf{99.86} \\
\midrule
\gptthreepar & \textbf{92.16} & \textbf{97.96} & \textbf{94.10} & \textbf{99.46} & 92.82 & \textbf{90.34} & \textbf{99.83} & \textbf{99.91} \\
\gptfourpar & 91.11 & 97.91 & 93.74 & 99.36 & \textbf{95.55} & 88.94 & 99.83 & 99.77 \\

\bottomrule
\end{tabular}
% }
\smallskip
\caption{\textbf{Ablation Study of LLM Architecture.} \llmassist evaluated on nuPlan Closed-Loop Reactive Challenge on Val14 split. GPT-3 and GPT-4 achieve comparable performance when used as the LLM for \llmassist.}
\label{tab:gpt3vs4}
\vspace{-1em}
\end{table*}

%% file: sec/5_conc.tex
\section{Discussions}
\label{sec:discussions}
    In the realm of autonomous driving, the potential of language models, particularly in robotic tasks, has been increasingly recognized. The introduction of Large Language Models (LLMs) for complex reasoning tasks, as highlighted in works like those by \citet{brohan2023rt} and \citet{ahn2022can}, and their application in agentic control as demonstrated by \citet{wang2023voyager}, marks a significant advancement in this field. This trend is expected to amplify as foundation models grow larger, enhancing the emergent abilities of LLMs for more effective reasoning and grounding in real-world scenarios.

    However, there are notable limitations to our current approach. We rely on a text-only model that processes a parsed state, which, while abstract, cannot fully replace perception-based systems in terms of information richness and context. Additionally, the speed at which decisions must be made in autonomous driving poses a challenge. Current LLMs operate slower than required for time-sensitive decision-making. While our model does not directly address these speed constraints, we anticipate future advancements in LLMs to bring improvements in both capability and processing speed. Another critical concern is the tendency of LLMs to produce hallucinated outputs \citep{mckenna2023sources, mundler2023self}. Current directions of research such as certified reasoning \citep{poesia2023certified} and retrieval feedback \citep{yu2023improving} seek to mitigate this, but significant work remains, especially in high-risk domains like autonomous driving where accuracy is paramount and human lives are at stake.
    
    Looking forward, our research underscores the promising role of LLMs in enhancing the performance of rule-based planners in scenarios where they fall short. We posit that future research should focus on rectifying the existing flaws of LLMs. This includes improving their grounding, incorporating multiple modalities for richer contextual understanding, and enhancing their scalability and speed. Such advancements will not only bolster the efficacy of LLM-assisted planning systems but also expand the applicability of LLMs in various facets of autonomous navigation and beyond.

\section{Conclusion}
\label{sec:conclusion}
    In this paper, we introduce \llmassist, a novel approach in closed-loop planning for autonomous driving that synergizes the advanced capabilities of Large Language Models (LLMs) with traditional rule-based methods. Harnessing the emergent commonsense reasoning nad cognitive abilities of recent LLMs, particularly GPT 3/4, \llmassist excels in navigating intricate driving scenarios where conventional methods are often insufficient. Our comprehensive evaluations on the nuPlan benchmark confirm \llmassist's state-of-the-art performance in both reactive and non-reactive settings across numerous metrics of driving. Our work not only demonstrates the efficacy of integrating lanuage models into autonomous driving solutions but also underscores their significance and ability in refining and enhancing complex decision-making processes, setting a new benchmark for future developments in this field.

%% file: sec/6_suppl.tex
\clearpage
\setcounter{page}{1}

\twocolumn[{%
\renewcommand\twocolumn[1][]{#1}%
\maketitlesupplementary
\begin{center}
\input{tbl/queries}
\end{center}
\vspace{1em}
}]

\section{Ablation Study of Number of LLM Queries}
\label{sec:queries}
\begin{figure*}
    \centering
    \includegraphics[width=1\linewidth]{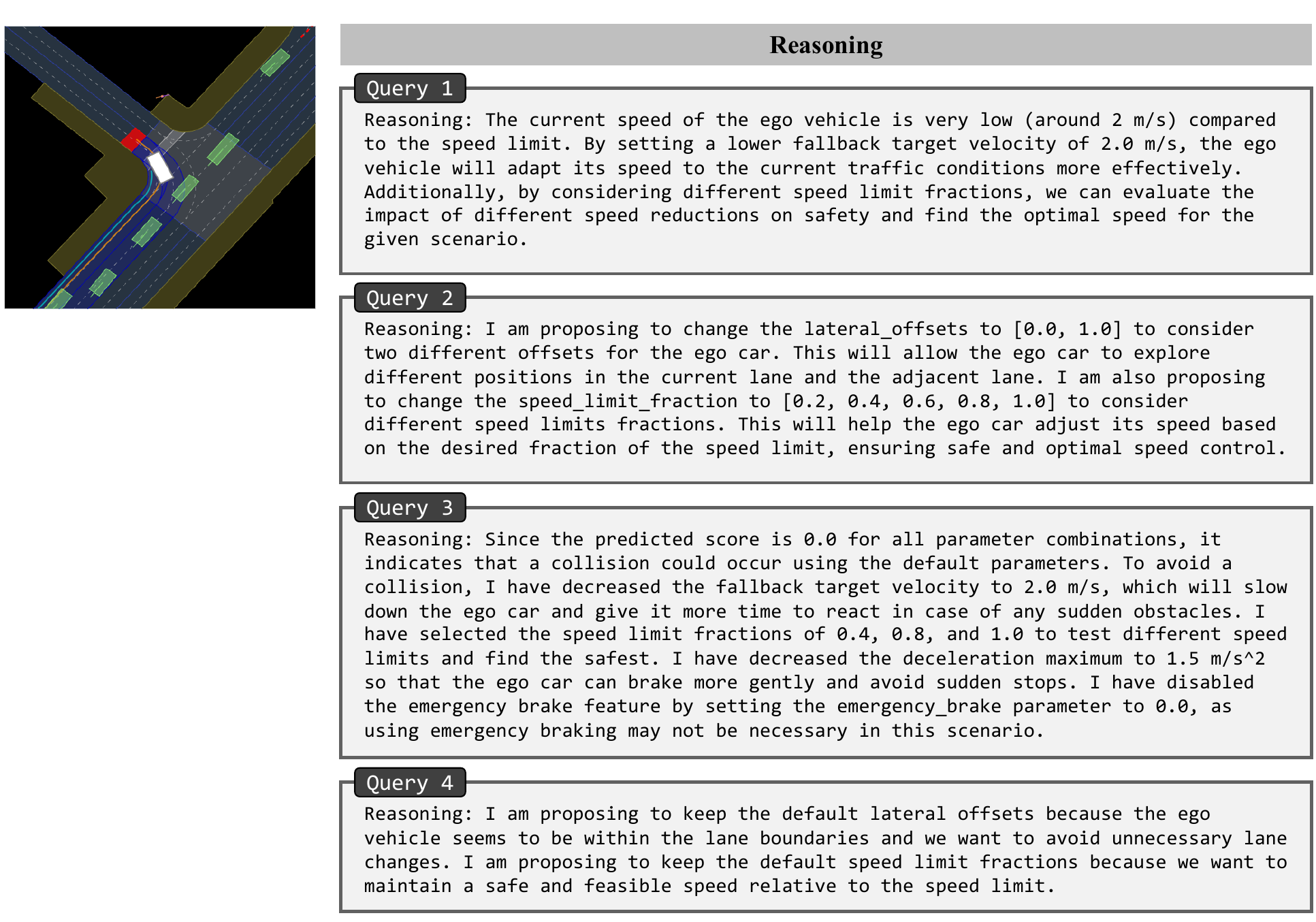}
    \caption{\textbf{Qualitative Example of Making Hard Turns by Slowing Down by \llmassist (4 Query Non-Reactive).} In another challenging scenario, the PDM Closed system encountered difficulties, colliding with traffic in the adjacent left lane due to maintaining a high speed during a turn, which resulted in a larger turning radius. However, as depicted above, \llmassist approaches this situation with a more nuanced strategy. Initially recognizing the low velocity, it experiments with different speed fractions. Through iterative feedback from multiple rounds of prompting, it discerns that maintaining a slow speed is crucial to avoid collisions and achieve higher performance scores. Consequently, \llmassist opts to stay centered in the lane, completing the turn at a reduced speed and successfully navigating the scenario without incident. This illustrates \llmassist's ability to learn from feedback, adapt its strategy accordingly, and execute precise maneuvers under complex driving conditions.}
    \label{fig_qual_4qnr}
\end{figure*}

\begin{figure*}
    \centering
    \includegraphics[width=1\linewidth]{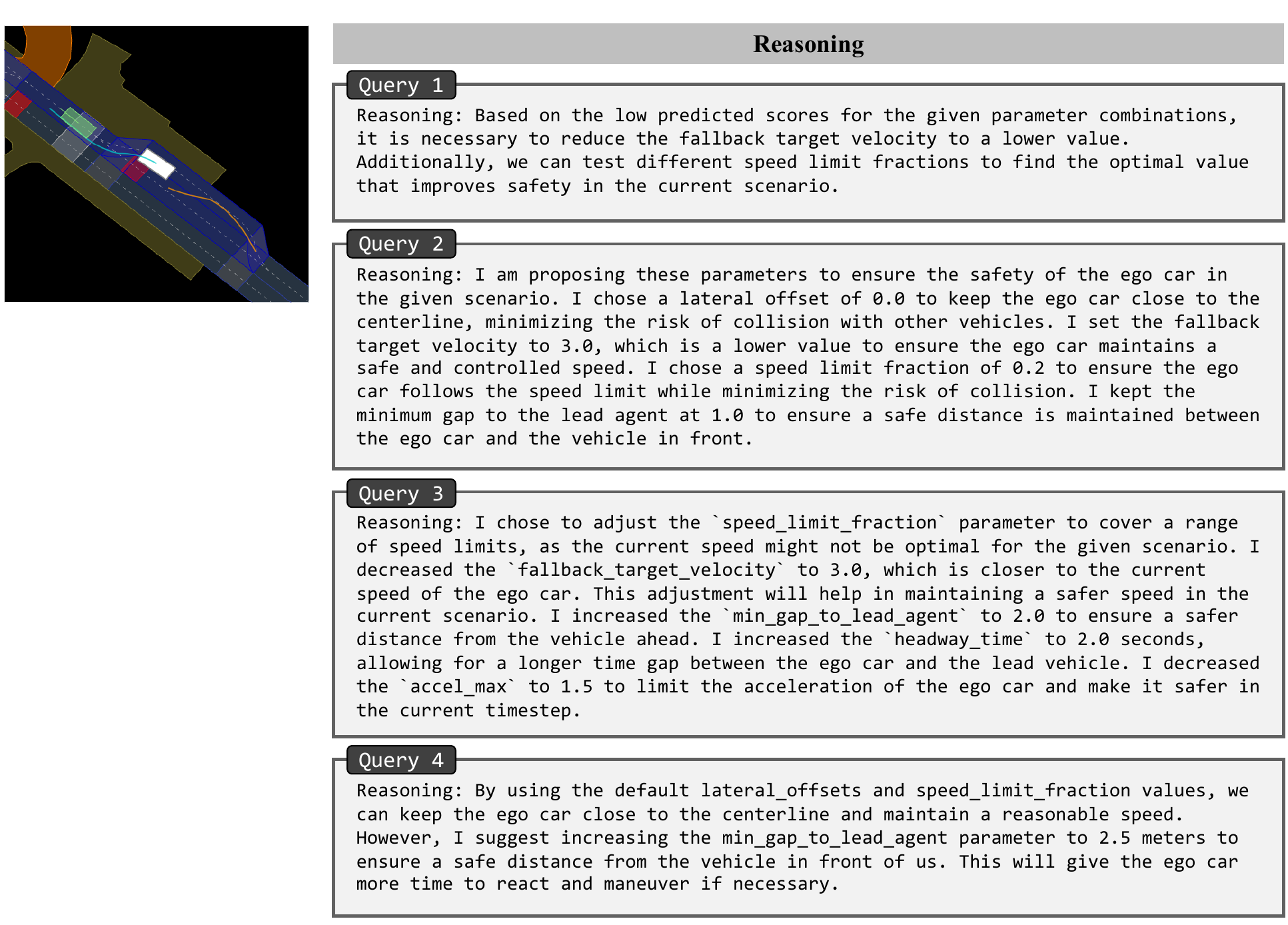}
    \caption{\textbf{Qualitative Example of Complex Maneuver by \llmassist (4 Query Reactive).} Now we examine a challenging scenario where the lane unexpectedly narrows, requiring proactive maneuvering. In this situation, it's crucial for the vehicle to initiate a leftward movement early, rather than waiting for the lane to curve. The PDM Closed system, constrained by its limited functionality, fails to execute this complex maneuver, resulting in the vehicle driving off-road and receiving a low score. In stark contrast, \llmassist, informed by four rounds of iterative prompting, quickly recognizes the need for a careful approach. As shown above, it calculates that the ego vehicle must reduce speed and change lanes in time to successfully merge. This successful execution of a demanding maneuver demonstrates \llmassist's advanced capability to adaptively respond to intricate driving scenarios, ensuring safe and efficient navigation.}
    \label{fig_qual_4qr}
\end{figure*}

As illustrated in \cref{fig:overview}, \llmassist queries the LLM planner multiple times at a given time step until a trajectory is proposed that has a predicted score that meets a predefined threshold or the number of queries per time step exceeds a predefined threshold. If the query threshold is exceeded, the trajectory with the highest predicted score is selected. In \cref{tab:num_rec}, we vary the number of allowed queries per time step and report the results. Note, since \llmassist uses PDMClosed \cite{Dauner2023CORL}, the current SoTA, as the base planner, the rows with 0 queries denote PDMClosed. The results show a clear trend -- as the number of LLM queries increases, the performance of \llmassist improves. We also show some qualitative results of multiple queries in practice in \Cref{fig_qual_4qnr} and \Cref{fig_qual_4qr}.

\section{Qualitative Results}
\begin{figure*}
    \centering
    \includegraphics[width=1\linewidth]{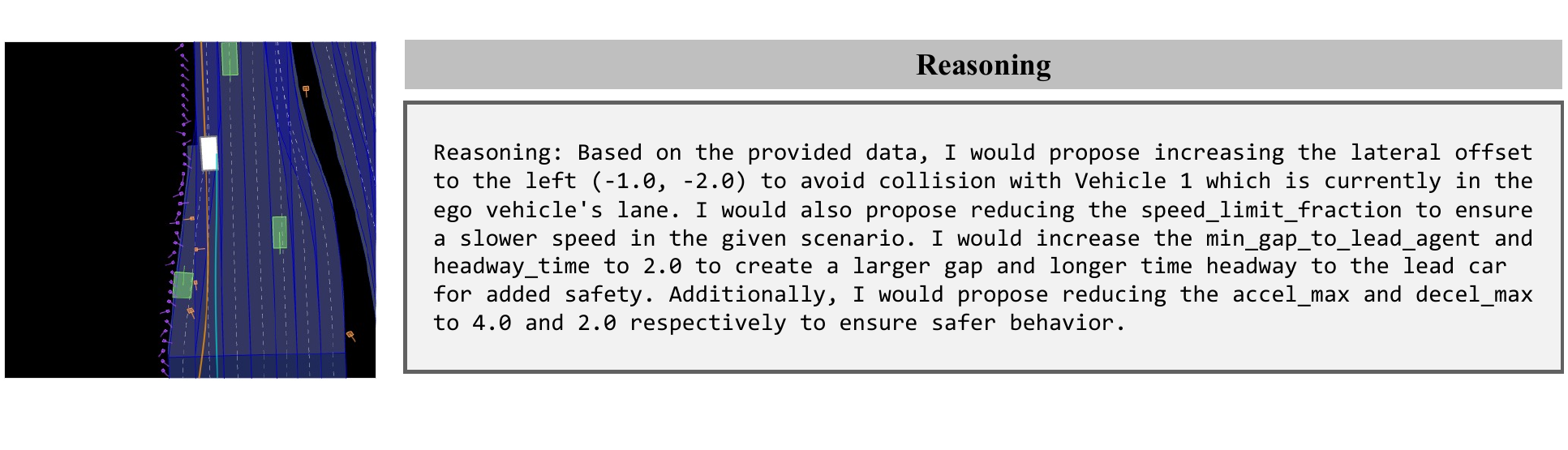}
    \caption{\textbf{Qualitative Example of Collision Aversion by Lane Changing by \llmassist (1 Query Non-Reactive).} In one notable scenario, the original PDM Closed planner exhibited a tendency to drive perilously close to traffic cones positioned on the left side of the lane, resulting in a low score. However, as illustrated above, our LLM planner adopts a more cautious strategy. It intelligently decides to swerve leftward and simultaneously reduce speed to safely navigate past the obstacles. This maneuver is visually represented in the figure, where the new plan trajectory, marked in cyan blue, effectively avoids the cones depicted as orange squares. This scenario underscores \llmassist's ability to accurately interpret the scene and adaptively change lanes when necessary to prevent collisions, highlighting its advanced scene understanding and responsive decision-making capabilities.}
    \label{fig_qual_1qnr}
\end{figure*}

\begin{figure*}
    \centering
    \includegraphics[width=1\linewidth]{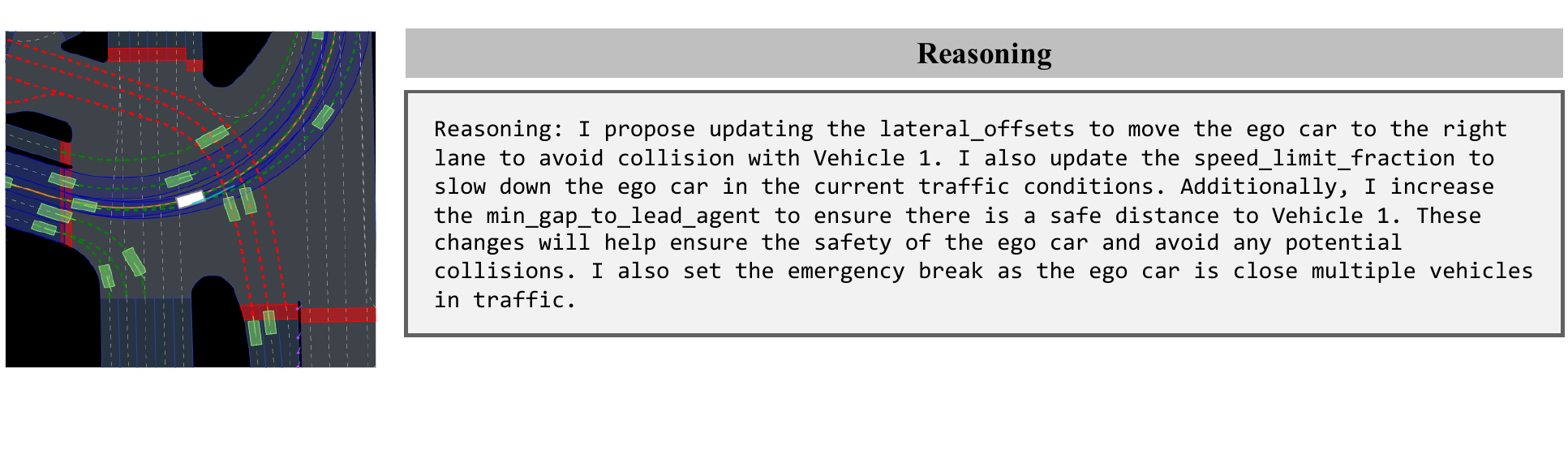}
    \caption{\textbf{Qualitative Example of Emergency Stopping by \llmassist (1 Query Reactive).} In a particularly intriguing scenario, the ego vehicle encounters an unexpected traffic challenge: while it is driving in its lane, traffic in a perpendicular lane suddenly starts moving due to a green light, an anomaly in typical traffic light operation. The PDM Closed system, unable to comprehend the movement of unrelated traffic, unfortunately collides with it. In contrast, as demonstrated above, \llmassist adeptly detects this unusual traffic behavior. It responds by slowing down and executing a swerve maneuver to sidestep a potential collision. Furthermore, \llmassist proactively invokes the emergency brake as an additional preventive measure. This instance vividly illustrates \llmassist’s superior capability to recognize and react to complex, dynamic traffic situations, ensuring safety through quick and effective decision-making.}
    \label{fig_qual_1qr}
\end{figure*}

In this section, we delve deeper into an array of qualitative outcomes, akin to those illustrated in \Cref{fig_qual}, to further demonstrate the efficacy of our method. This includes a selection of results from the multiple rounds of queries, a novel aspect introduced in this supplementary section \Cref{sec:queries}. Specifically, through \Cref{fig_qual_1qnr}, \Cref{fig_qual_4qnr}, \Cref{fig_qual_1qr}, and \Cref{fig_qual_4qr}, we showcase a diverse array of scenarios by selecting examples from each permutation of 1 versus 4 queries and Reactive versus Non-Reactive settings. This approach provides a comprehensive view of the versatility and robustness of our method across different driving conditions. 

\section{Metric Definitions}

For all metrics, we use the official nuPlan challenge \cite{caesar2021nuplan} definitions. We paraphrase these definitions below.

\paragraph{Score.} For each scenario, a combined score for the driven trajectory is calculated using a hybrid hierarchical-weighted average of individual metric scores. The planner receives a zero score in scenarios where (a) an at-fault collision with a vehicle, pedestrian, or bicyclist occurs, (b) multiple at-fault collisions with objects (like cones) happen, (c) there's a drivable area violation, (d) the ego vehicle enters oncoming traffic by more than 6 meters, or (e) insufficient progress is made by the ego. If there's a single at-fault collision with an object, or the ego drives into oncoming traffic for more than 2 meters but less than 6 meters, the weighted average score of other metrics is halved. In all other cases, the score is simply the weighted average of other metrics.

\paragraph{Collisions.} A collision occurs when the bounding box of the ego vehicle intersects with the bounding box of another agent. Regardless of the duration of the collision, it is counted as one event, and the initial frame is used to determine the kinetic energy transferred during the collision. Following the collision, any tracks involved are excluded from metric assessments in subsequent frames.

\paragraph{Time-to-Collision.} TTC, or Time-to-Collision, estimates the time until the ego vehicle potentially collides with another track, based on their current trajectories. It's calculated for tracks ahead, in cross traffic, or at the sides, particularly when the ego is changing lanes or in an intersection. TTC is determined by projecting the bounding boxes of the ego and other tracks forward at 0.1-second intervals, up to 3 seconds. The TTC is the earliest intersection time of these projections; if there's no intersection, the TTC is deemed infinite.

\paragraph{Drivable.} The drivable area compliance metric tracks instances where the ego veers outside this area. A small deviation outside the drivable area is permissible due to the overestimation of the ego's bounding box, with a maximum violation threshold of 0.3 meters. If any frame shows the ego's bounding box corners exceeding this threshold distance from the nearest drivable area, the compliance score is reduced to 0; otherwise, it remains at 1.

\paragraph{Comfort.} The comfort of the ego vehicle's trajectory is assessed by comparing key variables—minimum and maximum longitudinal accelerations, maximum lateral acceleration, yaw rate, yaw acceleration, longitudinal jerk, and overall jerk magnitude—to predefined thresholds. These thresholds are empirically set based on expert trajectory data (e.g., max longitudinal acceleration at 2.40 m/s$^2$, max lateral acceleration at 4.89 m/s$^2$).

\paragraph{Progress.}  To measure the ego vehicle's progress in a scenario, it is compared to an expert driver's route progress. This involves calculating the ego's progress per frame along the same lanes and lane connectors used by the expert, summing this progress over the scenario. The ego-to-expert progress ratio is then derived by comparing the ego's total progress to that of the expert. If the ego's progress falls below a certain negative threshold (-0.1m) due to data noise, the ratio is set to 0. In cases where no expert route is defined, the ratio defaults to 1. Otherwise, the ratio is the minimum of 1 and the adjusted ego-to-expert progress comparison. 

\paragraph{Speed Limit.} This metric checks if the ego vehicle's speed surpasses the speed limit, which is determined from the lane or, for a lane connector, from the higher speed limit of connecting lanes. A speed limit violation is noted whenever the ego's speed exceeds this limit.

\paragraph{Direction.} This metric is designed to penalize the ego vehicle if it enters oncoming traffic lanes. It calculates the movement of the ego's center over a 1-second period in relation to the designated driving direction, based on the baselines of the lanes or lane-connectors associated with the ego. The score is assigned a value of 1 if the ego does not travel against the traffic flow by more than 2 meters. If the ego moves against the traffic flow exceeding 6 meters, the score is set to 0. For movements between these two thresholds, the score is adjusted to 0.5.

%% file: tbl/queries.tex
% \begin{table*}[t]
\centering
\setlength{\aboverulesep}{0pt}
\setlength{\belowrulesep}{0pt}
\setlength{\extrarowheight}{.75ex}
\resizebox{0.9\textwidth}{!}{%
\begin{tabular}{c >{\columncolor{gray!50}}c >{\columncolor{gray!15}}c cccccccccc}
\toprule
\textbf{Challenge} & \textbf{Queries} & \textbf{Score} & \textbf{Collisions} & \textbf{TTC}  & \textbf{Drivable}  & \textbf{Comfort}  & \textbf{Progress}  & \textbf{Speed Limit}  & \textbf{Direction} \\
\midrule
\multirow{4}{*}{\shortstack{Closed-Loop\\Non-Reactive}} 
& 0 & 92.51 & 98.05 & 93.11 & \textbf{99.55} & 95.19 & 91.75 & \textbf{99.83} & \textbf{99.95} \\
& 1 & 92.52 & 98.32 & 92.92 & \textbf{99.55} & \textbf{95.74} & 91.16 & \textbf{99.83} & \textbf{99.95} \\
& 2 & 92.82 & \textbf{98.41} & 93.47 & 99.27 & 95.10 & 92.12 & \textbf{99.83} & \textbf{99.95} \\
& 4 & \textbf{93.05} & 98.31 & \textbf{93.69} & 99.54 & 95.61 & \textbf{92.16} & \textbf{99.83} & \textbf{99.95} \\
\midrule
\multirow{4}{*}{\shortstack{Closed-Loop\\Reactive}} 
& 0 & 91.79 & 97.91 & 93.29 & 99.37 & 94.65 & 89.92 & \textbf{99.83} & \textbf{99.95} \\
& 1 & 92.16 & 97.96 & \textbf{94.10} & 99.46 & 92.82 & \textbf{90.34} & \textbf{99.83} & 99.91 \\
& 2 & 92.18 & 98.00 & 93.81 & 99.45 &\textbf{95.36} & 90.10 & \textbf{99.83} & \textbf{99.95} \\
& 4 & \textbf{92.20} & \textbf{98.18} & 93.62 & \textbf{99.64} & 94.72 & 90.07 & \textbf{99.83} & \textbf{99.95} \\
\bottomrule
\end{tabular}
}
\smallskip
\captionof{table}{\textbf{Ablation Study of Number of LLM Queries per Iteration}. \gptthreepar evaluated on nuPlan Closed-Loop Challenges on Val14 split. Note, the rows with 0 queries denote PDMClosed \cite{Dauner2023CORL}, the current SoTA.}
\label{tab:num_rec}
% \end{table*}